\pdfoutput=1
\documentclass[11pt]{article}

\usepackage[preprint]{acl}

\usepackage{algorithmicx}
\usepackage{algpseudocode}
\usepackage{algorithm}
\usepackage{times}
\usepackage{latexsym}
\usepackage[utf8]{inputenc}
\usepackage{listings}

\usepackage[T1]{fontenc}
\usepackage[utf8]{inputenc}

\usepackage{microtype}

\usepackage{inconsolata}

\usepackage{graphicx}
\usepackage{geometry}
\usepackage{amsmath}
\title{A Multi-Model Adaptation of Speculative Decoding for Classification}

\author{
  \textbf{Somnath Roy},
  \textbf{Padharthi Sreekar},
  \textbf{Srivatsa Narasimha},
  \textbf{Anubhav Anand}
\\
  Freshworks Inc, USA
\\
  \small{
   \href{mailto:email@domain}{\{somnath.roy, venkatasai.sreekar, srivatsa.narasimha, anubhav.anand\}@freshworks.com}
}
}

\begin{document}
\maketitle
\begin{abstract}
The current study introduces a novel adaptation of speculative decoding, repurposed from generation to classification tasks. We propose a multi-model framework employing up to three lightweight "worker" models and a single, more robust "judge" model analogous to draft models and target model, respectively, in speculative decoding. The worker models, tasked with the bulk of the computation, independently predict discrete class labels for a given input. When majority worker models agree on a label, it is accepted as the final label, optimizing efficiency by bypassing the computationally expensive judge model. In cases of disagreement, the judge model intervenes to resolve the label. This approach minimizes redundant computation, leverages the redundancy of multiple workers for confidence, and confines the judge model's role to challenging cases, offering a practical balance of efficiency and accuracy. Our analysis suggests that smaller out of the box instruction/chat finetuned worker models with 3 billion parameters (hereafter, 3B) demonstrate a level of alignment with judge models comparable to that of larger finetuned worker models with 7 billion parameters (hereafter, 7B) across both simple and higher order reasoning tasks. The top performing 3B worker model pair achieve an agreement rate of approximately 80–83\% for sentiment and around 50–80\% for similar ticket when compared to judge models. Additionally, 3B worker models provide a speedup ranging from 2.8x to 9x relative to the judge models, while 7B worker model combinations achieve a speedup ranging from 1.28x to 0.28x. 
\end{abstract}

\section{Introduction}
Human annotators can easily adhere to instruction and style, but evaluating factual and logical accuracy grows harder with complex queries. This can lead annotator to favor plausible or longer responses, prioritizing style over accuracy. In other words, human annotation becomes unreliable especially for high order reasoning tasks or difficult tasks in general \cite{tan2024judgebench}.  Due to this reason Large Language Models (LLMs) are increasingly being utilized as automated evaluators to assess the quality and correctness of generated content, offering a scalable alternative to traditional human evaluation \cite{zheng2023judging}. Recent studies demonstrate that LLMs achieve over 80\% agreement with human preferences across various evaluation tasks, underscoring their reliability as judges~\cite{bai2023benchmarking}. Fine-tuning further enhances their accuracy, particularly in domain-specific applications such as dialogue evaluation \cite{zhu2023judgelm}. Moreover, the increasing availability of highly capable open-source LLMs of different sizes, as well as smaller reasoning models, presents new opportunities. These models, often trained on datasets comparable to those of advanced closed-source LLMs, offer a cost-effective alternative for evaluation tasks. It is important to note that it's deployment as independent judge can be unreliable, particularly in crucial tasks that influence business-critical decisions. Also with the growing demand of agentic systems where LLMs generate responses and make critical decisions, it is essential to ensure the reliability at each step at a very low latency. Therefore, A lightweight framework is required to evaluate the accuracy of both their outputs. These evaluation tasks can be formulated as a classification problem. Inspired by speculative decoding, a technique originally developed to accelerate token generation, we propose adapting this approach for classification. The following is our novel contribution.
\begin{itemize}
    \item Our framework leverages multiple lightweight "worker" models to predict discrete class labels, reserving a robust "judge" model for cases of disagreement. This method optimizes efficiency by minimizing the judge model's involvement while harnessing the diversity of smaller models to enhance reliability.
    \item Unlike speculative decoding, we ensure that all the models employed in our framework, including from workers to judge, are non-overlapping in terms of their training architecture and instruction finetuning datasets. Diversity in models helps minimize correlated errors. 
    \item Rather than relying solely on disagreement cases between individual workers or worker combinations, we propose that incorporating a confidence threshold at the worker level can further enhance fidelity.
    \end{itemize}

 In this study, we explore this multi-model strategy, addressing both scalability of LLM evaluation and the practical constraints of existing paradigms. 

\section{Background}
A decoder-only LLM performs inference in two phases: prefill and autoregressive decoding \cite{yan2024decoding}. In the prefill phase, it processes the entire input context in parallel for next-word prediction. In autoregressive decoding, it generates tokens sequentially, making it memory bandwidth-bound on modern GPUs \cite{leviathan2023fast}. Speculative decoding leverages a smaller, faster draft model to generate token predictions, as autoregressive decoding latency increases linearly with layer depth despite having similar total parameters \cite{yan2024decoding}. These predictions are then verified or corrected by a bigger target model. The number of tokens accepted from the model draft output inherently depends on how well the model draft aligns with the distribution of the target model. If the predictions of the draft model closely match what the target model would produce, more tokens are accepted per iteration, thereby accelerating generation. However, if the alignment is poor, fewer tokens are accepted and the process leans more towards the target model, reducing efficiency \cite{liu2023online}. The choice of a draft model introduces variability because different draft models that vary in size, architecture, or training data will produce different results, leading to fluctuations in how many tokens are accepted before rejection. Even factors like temperature or the way the draft model was distilled from the target model can affect the alignment, making the outcome less predictable. Therefore, the overall algorithm follows a deterministic structure, i.e., fixed models and inputs, however, the interplay between draft and target introduces a non-determinism in performance, tied directly to that model alignment.\\

Consider a judge model \( J \) generating tokens \( x_1, x_2, \dots, x_n \) with probability distribution \( P_J(x_t | x_{<t}) \), and a worker model \( W \) proposing \( k \) speculative tokens \( y_1, y_2, \dots, y_k \) with distribution \( P_W(y_t | y_{<t}, x_{<t}) \).

\subsubsection*{Acceptance Criterion}
The judge model accepts a token \( y_t \) if its probability aligns with the worker's prediction within a random threshold \( r_t \sim \text{Uniform}[0, 1] \):
\begin{equation}
\frac{P_J(y_t | x_{<t}, y_{<t})}{P_W(y_t | x_{<t}, y_{<t})} \geq r_t
\end{equation}

\subsubsection*{Number of Accepted Tokens}
The number of accepted tokens \( m \) (where \( 0 \leq m \leq k \)) is defined as:
\begin{equation}
m = \max \left\{ t \mid \forall j \leq t, \frac{P_J(y_j | x_{<t}, y_{<j})}{P_W(y_j | x_{<t}, y_{<j})} \geq r_j \right\}
\end{equation}

\subsubsection*{Distribution Alignment Limitation}
The divergence between the judge and worker distributions is measured by the KL-divergence:
\begin{equation}
D_{KL}(P_J || P_W) = \sum_{y} P_J(y | x_{<t}) \log \left( \frac{P_J(y | x_{<t})}{P_W(y | x_{<t})} \right)
\end{equation}
Perfect alignment (\( D_{KL}(P_J || P_W) = 0 \)) implies \( P_W = P_J \), but for a smaller worker model (\( |W| < |J| \)):
\begin{equation}
\min_{D} D_{KL}(P_J || P_W) > 0
\end{equation}
The worker's distribution approximates the judge's with an error term:
\begin{equation}
P_W(y | x_{<t}) \approx P_J(y | x_{<t}) + \epsilon(y, x_{<t})
\end{equation}

\subsubsection*{Non-Determinism}
The variance in the number of accepted tokens reflects the non-deterministic nature:
\begin{equation}
\text{Var}(m) = \text{Var} \left( \max \left\{ t \mid \text{alignment holds} \right\} \right)
\end{equation}

\section{Extending Speculative Decoding for Classification}

We propose speculative decoding for classification, using up to three worker models to predict discrete class labels, with a judge model only predicting the disagreed samples among workers. For example: an input \( x \) and discrete classes \( C = \{c_1, c_2, \dots, c_m\} \), define worker models \( W_1, W_2, W_3 \), each predicting a label \( y_i = W_i(x) \in C \) with confidence \( p_i(y_i | x) \), and a target model \( T \) predicting \( y_T = T(x) \).

\subsection*{Workers Prediction}
\begin{equation}
y_i = W_i(x), \, p_i(y_i | x), \, i = 1, 2, 3
\end{equation}

\subsection*{Agreement Rules}
Many agreement rules can be defined based on the number of workers and judges involved. However, the following three are the most prominent and commonly utilized.
\subsubsection*{Criterion 1: Simple Majority}
Let \( n(y) = |\{i : y_i = y\}| \). Then:
\begin{equation}
y_{\text{draft}}^1 = 
\begin{cases} 
y & \text{if } \exists y : n(y) \geq 2, \\
\text{undefined} & \text{otherwise}
\end{cases}
\end{equation}

\subsubsection*{Criterion 2: Majority with Confidence Threshold}
Let \( n(y) = |\{i : y_i = y\}| \), \( P(y) = \{p_i : y_i = y\} \), and \( \sigma^2 = \text{Var}(\{p_1, p_2, p_3\}) \). Then:
\begin{equation}
y_{\text{draft}}^2 = 
\begin{cases} 
y & \text{if } \exists y : n(y) \geq 2, \\
  & \quad \min(P(y)) \geq \tau, \\
  & \quad \sigma^2 \geq \delta \\
\text{undefined} & \text{otherwise}
\end{cases}
\end{equation}

\subsubsection*{Criterion 3: Unanimous}
\begin{equation}
y_{\text{draft}}^3 = 
\begin{cases} 
y_1 & \text{if  $y_1$ = $y_2$ = $y_3$} \\
\text{undefined} & \text{otherwise}
\end{cases}
\end{equation}
The final label can be decided based on any of the above criterion as shown below. 
\subsection*{Final Label Selection}
\begin{equation}
y_{\text{final}} = 
\begin{cases} 
y_{\text{draft}}^1 \space \text or \text \space y_{\text{draft}}^2 \text or \text \space y_{\text{draft}}^3 & \text{if defined} \\
T(x) & \text{otherwise}
\end{cases}
\end{equation}

Current work exploits criterion 2 and the worker number and number of label conflicts is handled using following using Algorithm 1. 

\begin{algorithm}
\caption{Majority Criterion with Grouping}
\begin{algorithmic}[1]
\State \textbf{Input:} Set of worker annotations $\{y_i\}_{i=1}^{W}$, number of labels $L$
\State \textbf{Output:} best agreement worker pair

\If{$W \leq L$}
\State Compute $n(y) = |{i : y_i = y}|$ for each label $y$.
\If{there exists $y$ such that $n(y) \geq 2$}
\State Set $y_{\text{draft}}^1 = y$.
\Else
\State Set $y_{\text{draft}}^1$ as undefined.
\EndIf
\Else
\State Compute all possible worker groupings as $\binom{W}{L}$.
\State Find the label pair $(y_a, y_b)$ having agreement for each worker pair.
\State Return worker pair with highest agreement.
\EndIf

\end{algorithmic}
\end{algorithm}

\section{Experiment}
There exists open-source benchmarks that includes datasets from various domain like psychology, biology, mathematics etc. \cite{tan2024judgebench, cobbe2021training, li2024generation}. Although several high-quality benchmarks have been introduced, concerns about their appropriate use and the fair comparison of different models continue to grow \cite{zhou2023don}. Therefore, this experiment is conducted using our proprietary datasets as described below.

\subsection{Dataset}
We have developed two datasets for the current study: one consisting of 10,000 samples randomly selected ticket data over the past month from 5 domains Retail, Healthcare, Logistics, Technology and Education. Ticket data is inherently complex due to the iterative interactions between customers and agents until a resolution is reached. The average number of interactions is 20. A single ticket may express multiple sentiments; however, the current evaluation focuses on the overall customer sentiment. The second dataset consists of 1,000 samples, where each sample includes a query—defined as the first message in a ticket—and three relevant tickets retrieved using our similar ticket pipeline. These samples are sourced from the healthcare domain.

\subsection{Model Selection}
We have selected three out of the box models for both 3B and 7B categories for worker. These models are either instruction finetuned or chat model and can be prompted for the classification task. Two judge models are selected: one is a dense model with 32B parameters, and the other is a mixture of experts (MoE) model consisting of 8 experts, each with 7B parameters (8×7B). The details of the model can be found below. \\

\begin{table}[h]
    \centering
    \small % Reduce font size
    \renewcommand{\arraystretch}{1.2} % Adjust row height
    \setlength{\tabcolsep}{1.2pt} % Adjust column spacing
    \begin{tabular}{|l|c|c|c|c|}
        \hline
        \textbf{Parameter} & \textbf{Worker 1} & \textbf{Worker 2} & \textbf{Worker 3} & \textbf{Judge 1} \\
        \hline
        \#hidden\_layers & 24 & 36 & 32 & 64 \\
        \hline
        hidden\_size & 3840 & 2048 & 3072 & 5120 \\
        \hline
        \#attention\_heads & 32 & 16 & 32 & 40 \\
        \hline
        intermediate\_size & 10240 & 11008 & 8192 & 27648 \\
        \hline
        vocab\_size & 32000 & 151936 & 32064 & 152064 \\
        \hline
        max\_pos\_embed & 8192 & 32768 & 4096 & 32768 \\
        \hline
    \end{tabular}
    \caption{Comparison of Model Specifications for Workers and Judge. Worker1, Worker2 and Worker3 represent the 3B category model namely h2o-danube3.1-4b-chat, Qwen2.5-3B-Instruct, Phi-3-mini-4k-instruct. Judge1 is 32B model named QwQ-32B-Preview }
    \label{tab:model_specs}
\end{table}

\begin{table}[h]
    \centering
    \small % Reduce font size
    \renewcommand{\arraystretch}{1.2} % Adjust row height
    \setlength{\tabcolsep}{1.2pt} % Adjust column spacing
    \begin{tabular}{|l|c|c|c|c|}
        \hline
        \textbf{Parameter} & \textbf{Worker 1} & \textbf{Worker 2} & \textbf{Worker 3} & \textbf{Judge2} \\
        \hline
        \#hidden\_layers & 32 & 28 & 32 & 32 \\
        \hline
        hidden\_size & 4096 & 3584 & 4096 & 4096 \\
        \hline
        \#attention\_heads & 32 & 28 & 32 & 32 \\
        \hline
        intermediate\_size & 14336 & 18944 & 14336 & 14336 \\
        \hline
        vocab\_size & 128256 & 152064 & 32768 & 32000 \\
        \hline
        max\_pos\_embed & 131072 & 131072 & 32768 & 32768 \\
        \hline
    \end{tabular}
    \caption{Comparison of Model Specifications for Workers and Judge2.  Worker1, Worker2 and Worker3 represent the 7B category model namely meta-Llama-3.1-8B-Instruct, DeepSeek-R1-Distill-Qwen-7B, Mistral-7B-Instruct-v0.3. Judge2 is 8x7B MOE model named Mixtral-8x7B-Instruct-v0.1. }
    \label{tab:model_specs}
\end{table}

For the sentiment analysis task, workers and judges are provided few-shot prompts to classify each ticket into one of three categories: positive, neutral, or negative. Similarly, for the similar ticket evaluation, workers and judges received few-shot prompts to assess the retrieved tickets for similarity, assigning a label of 1 for similar and 0 for not similar\footnote{Our findings indicate that these worker models frequently fail to adhere to the prescribed output format, particularly in reasoning tasks, due to the significantly higher number of max\_new\_tokens required. Furthermore, this hyperparameter varies across models, and when set to a lower value, some models may terminate prematurely without generating the actual class label.}. The proposed experimental setup incorporates two judges to evaluate potential biases toward the most closely related worker models used in this study. 

\subsection{Result}
 The agreement levels between each worker model and the judge model for both tasks are presented in Figure 1. Within the similar ticket category, h2odanube3.1-4b-chat exhibits the highest alignment with the judge model, whereas Qwen2.5-3B-Instruct demonstrates the lowest agreement. Similarly, within the 7B category, DeepSeek-R1-Distill-Qwen7B achieves the highest alignment with the judge model, while Mistral-7B-Instruct-v0.3 has the weakest performance in similar ticket identification. Conversely, across categories, Qwen2.5-3B-Instruct emerges as the best-performing model for sentiment analysis, with all other models displaying comparable but slightly lower agreement. These findings suggest that  worker models should selected on the basis of their performance specific to the task in hand.

As depicted in Figure 2, the agreement between the highest-performing worker models and the judge model ranges from 80-83\% for sentiment analysis and 46-80\% for similar ticket identification. This suggests that, for sentiment analysis, up to 20\% of inputs require review or labeling by the judge model. However, for similar ticket identification, 3B worker models necessitate judge model review or labeling in up to 54\% of cases, highlighting the increased complexity of this task.

The 95th percentile (P95) latency for both judge and worker models is illustrated in Figures 3 and 4. The top-performing 3B worker model combinations are approximately 9x and 2.18x faster compared to the judge model QwQ-32B-Preview and Mixtral-8x7B-Instruct-v0.1 respectively. While 7B worker models are only 1.19x faster to QwQ-32B-Preview and 2.88x slower compared to Mixtral-8x7B-Instruct-v0.1. Since 7B models achieve only a 3-5\% improvement in accuracy compared to their 3B counterparts, their substantially higher computational cost, being approximately 4 to 8 times slower, must be carefully considered. 

\section{Conclusion and Limitations}
Our findings suggest that carefully selected out-of-the-box 3B model combinations achieve accuracy comparable to that of 7B model combinations across both simple and higher-order reasoning tasks. Furthermore, 3B models, when used as workers, provide a substantial speedup of up to 9×, whereas 7B models do not offer significant speed advantages over the judge model. Additionally, we observe that out-of-the-box models exhibit considerable sensitivity to output style formatting in higher-order reasoning tasks. Therefore, some investment in prompt engineering experiment and hyperparameter tuning, particularly in optimizing max\_new\_tokens as well as post processing is required to achieve desired result for task in hand. Following are the limitations of the present study. 
\begin{itemize}
    \item This study includes only out-of-the-box, open-source, fine-tuned top-performing models from the 3B and 7B categories as worker models.
    \item The scope of this study is limited to Criterion 1, which is based on simple majority agreement. Future research should explore the application of Criterion 2.
    \item We observed that these out-of-the-box fine-tuned models generate a large number of tokens before producing the final output class, with significant variability across models. Consequently, future studies incorporating task-specific fine-tuned linear heads for classification could provide better control over output generation and facilitate the establishment of confidence thresholds. 
\end{itemize}

\begin{figure}[h]
    \centering
    \includegraphics[width=0.45\textwidth]{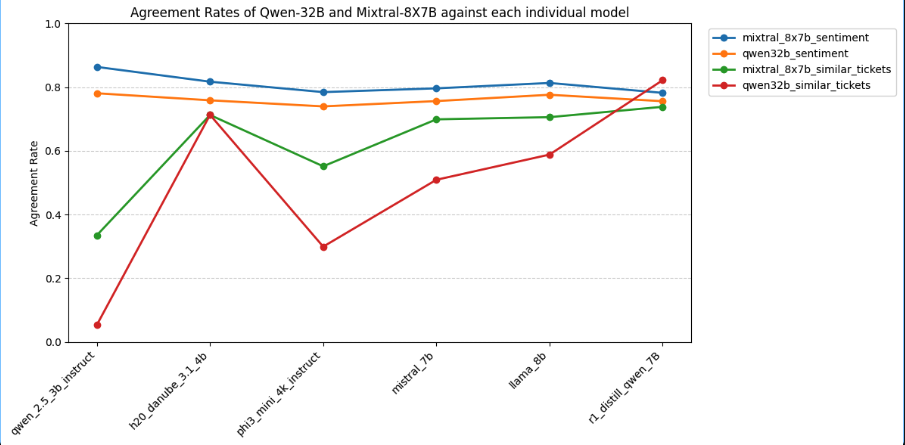} % Replace with your image file
    \caption{Agreement of worker models with each judge for sentiment and similar ticket task.}
    \label{fig:example_image}
\end{figure}

\begin{figure}[t]
    \centering
    \includegraphics[width=0.45\textwidth]{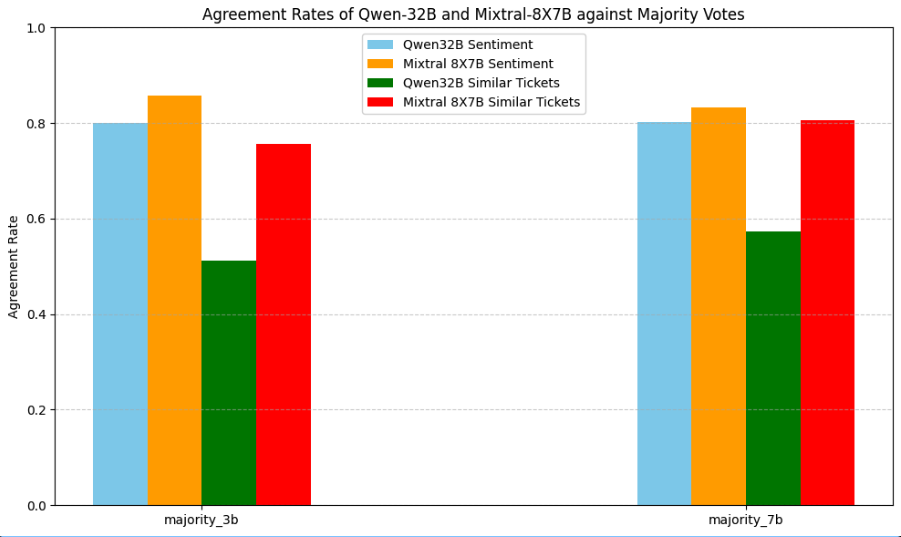} % Replace with your image file
    \caption{Agreement of best worker combinations with judge models for sentiment and similar ticket. The best worker pair for similar ticket is (Phi-3-mini-
4k-instruct, h2o-
danube3.1-4b-cha) in 3B category and (Mistral-7B-Instruct-v0.3, Llama-3.1-8B-Instruct) in 7B category.}
    \label{fig:example_image}
\end{figure}
\begin{figure}[t]
    \centering
    \includegraphics[width=0.45\textwidth]{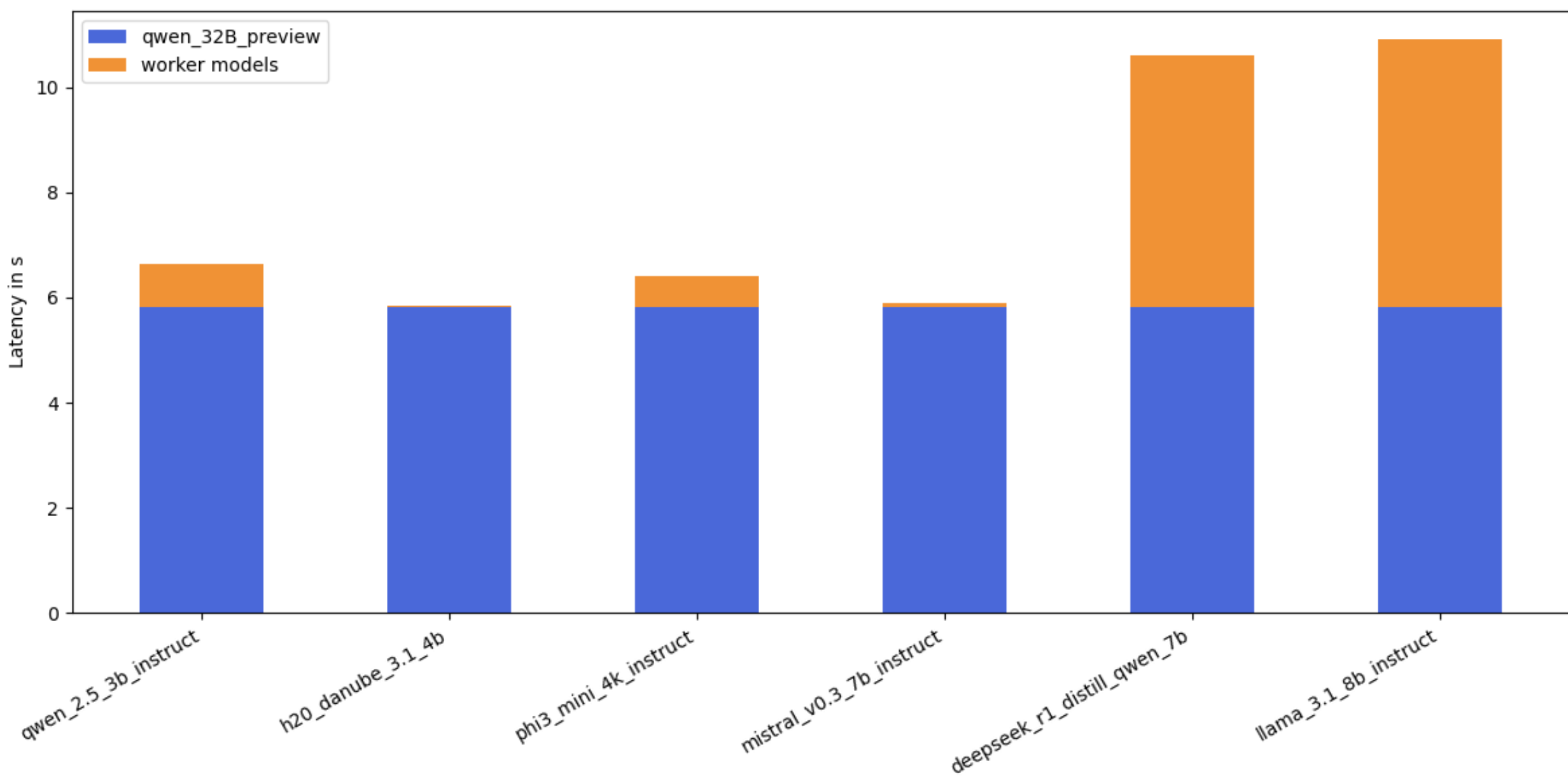} % Replace with your image file
    \caption{The p95 latency of the Qwen-32B-Preview judge model and the worker model was measured using vLLM with Flash Attention 1 on a single A100 GPU with 80GB VRAM}
    \label{fig:example_image}
\end{figure}

\begin{figure}[t]
    \centering
    \includegraphics[width=0.45\textwidth]{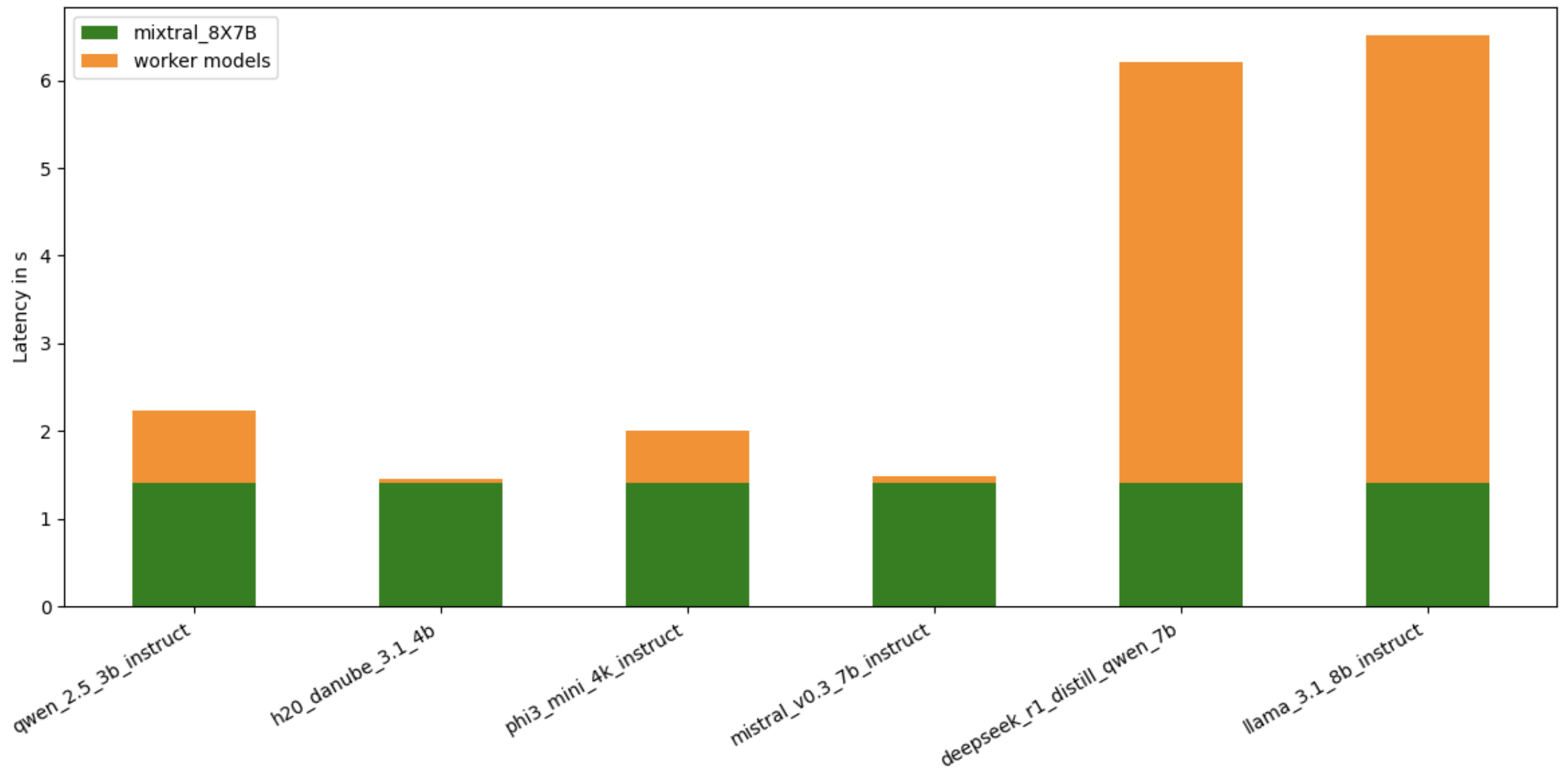} % Replace with your image file
    \caption{The p95 latency of the  Mixtral-8x7B-Instruct judge model and the worker model was measured using vLLM with Flash Attention 1 on a single A100 with 80GB RAM. }
    \label{fig:example_image}
\end{figure}
% Bibliography entries for the entire Anthology, followed by custom entries
%\bibliography{anthology}
% Custom bibliography entries only
\clearpage
 \bibliography{acl_latex}

\end{document}